\documentclass[sigconf,screen]{acmart}
\AtBeginDocument{%
  \providecommand\BibTeX{{%
    \normalfont B\kern-0.5em{\scshape i\kern-0.25em b}\kern-0.8em\TeX}}}

\usepackage{enumitem}

\newcommand{\red}[1]{\textcolor{black}{#1}}
\newcommand{\revision}[1]{\textcolor{black}{#1}}

\newcommand{\reffig}[1]{\textcolor{black}{Fig.~\ref{fig:#1}}}
\newcommand{\refsec}[1]{\textcolor{black}{Sec.~\ref{sec:#1}}}
\newcommand{\reftab}[1]{\textcolor{black}{Tab.~\ref{tab:#1}}}
\newcommand{\refeq}[1]{\textcolor{black}{Eq.~\ref{eq:#1}}}
\newcommand{\eg}[1]{{\textit{e.g.,~}}}
\newcommand{\ie}[1]{{\textit{i.e.,~}}}
\newcommand{\etal}[1]{{\textit{et al.~}}}

\copyrightyear{2024}
\acmYear{2024}
\setcopyright{acmlicensed}\acmConference[SIGGRAPH Conference Papers '24]{Special Interest Group on Computer Graphics and Interactive Techniques Conference Conference Papers '24}{July 27-August 1, 2024}{Denver, CO, USA}
\acmBooktitle{Special Interest Group on Computer Graphics and Interactive Techniques Conference Conference Papers '24 (SIGGRAPH Conference Papers '24), July 27-August 1, 2024, Denver, CO, USA}
\acmDOI{10.1145/3641519.3657471}
\acmISBN{979-8-4007-0525-0/24/07}

\begin{document}

% \title{InspireStation: Generating a Gallery of  3D Assets   from Few Examples}
\title{ThemeStation: Generating Theme-Aware 3D Assets from Few Exemplars}
%Few Examples}


\author{Zhenwei Wang}
\authornote{Work done when interning at Shanghai Artificial Intelligence Laboratory.}
\email{zhenwwang2-c@my.cityu.edu.hk}
\orcid{0000-0003-0215-660X}
\affiliation{%
  \institution{City University of Hong Kong}
  \city{Hong Kong SAR}
  \country{China}
}

\author{Tengfei Wang}
\authornote{Co-corresponding authors.}
\email{tfwang@connect.ust.hk}
\orcid{0000-0002-3435-8110}
\affiliation{
  \institution{Shanghai AI Lab}
  \city{Shanghai}
  \country{China}
}

\author{Gerhard Hancke}
\email{gp.hancke@cityu.edu.hk}
\orcid{0000-0002-2388-3542}
\affiliation{%
  \institution{City University of Hong Kong}
  \city{Hong Kong SAR}
  \country{China}
  }

\author{Ziwei Liu}
\email{ziwei.liu@ntu.edu.sg}
\orcid{0000-0002-4220-5958}
\affiliation{
  \institution{Nanyang Technological University}
  \country{Singapore}
}

\author{Rynson W.H. Lau}
\authornotemark[2]
\email{rynson.lau@cityu.edu.hk}
\orcid{0000-0002-8957-8129}
\affiliation{%
  \institution{City University of Hong Kong}
  \city{Hong Kong SAR}
  \country{China}
}


\begin{abstract}
Real-world applications often require a large gallery of 3D assets that share a consistent theme.
While remarkable advances have been made in
%2D diffusion models have made remarkable advances 
general 3D content creation from text or image, synthesizing customized 3D assets following the shared theme of input 3D exemplars remains an open and challenging problem. 
In this work, we present \textit{ThemeStation}, a novel approach for theme-aware 3D-to-3D generation. \textit{ThemeStation} synthesizes customized 3D assets based on given few exemplars with two goals: \textit{1) unity} for generating 3D assets that thematically align with the given exemplars and \textit{2) diversity} for generating 3D assets with a high degree of variations. To this end, we design a two-stage framework that draws a concept image first, followed by a reference-informed 3D modeling stage. We propose a novel dual score distillation (DSD) loss to jointly leverage priors from both the input exemplars and the synthesized concept image. Extensive experiments and a user study confirm that \textit{ThemeStation} surpasses prior works in producing diverse theme-aware 3D models with impressive quality. \textit{ThemeStation} also enables various applications such as controllable 3D-to-3D generation.

\end{abstract}



\begin{teaserfigure}
    \centering
  \includegraphics[width=0.95\textwidth]{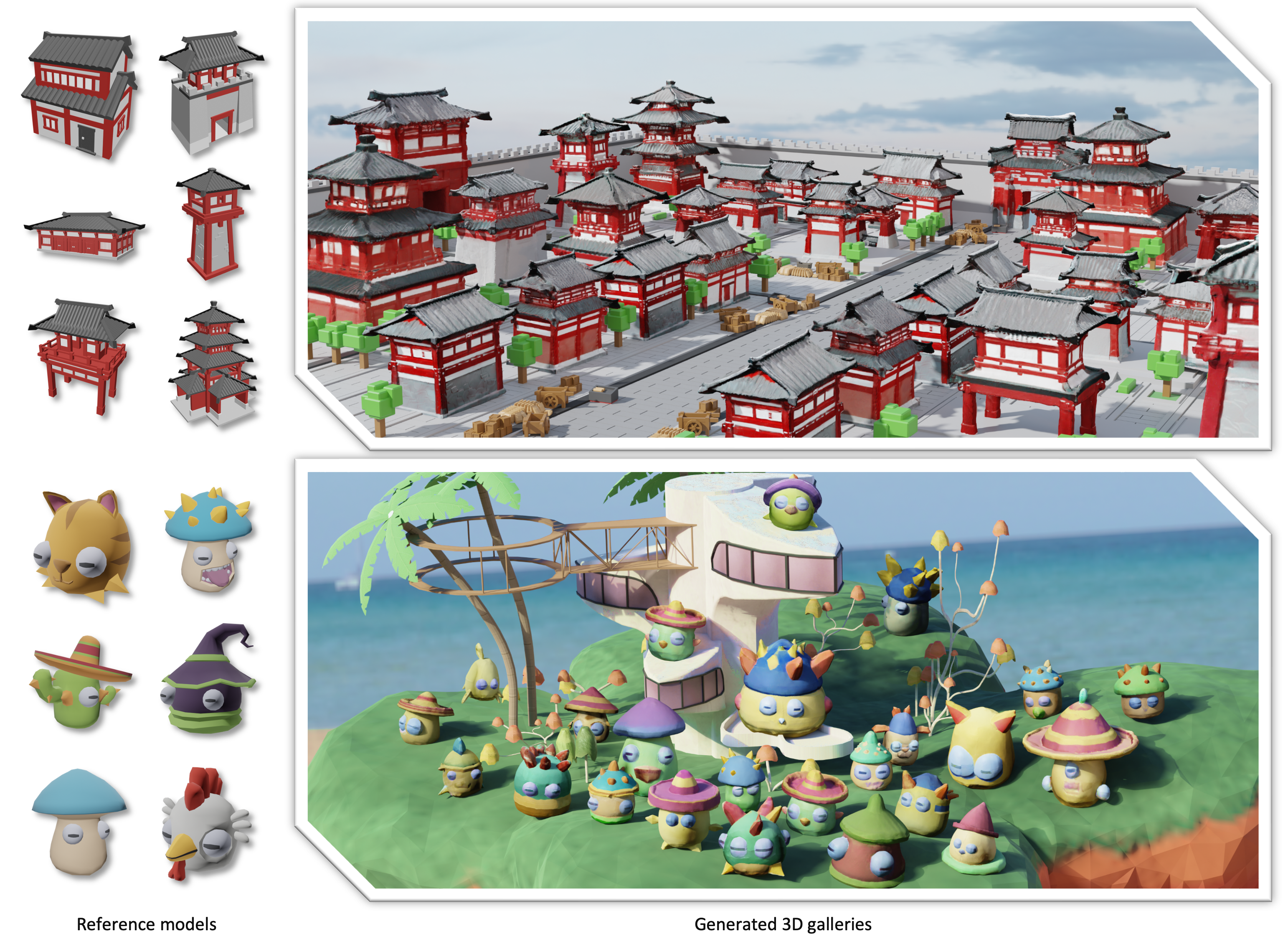}
  \vspace{-3mm}
  \caption{\textit{ThemeStation} can generate a gallery of 3D assets (right) from just one or a few exemplars (left). The synthesized models share consistent themes with the reference models, showing the immense potential of our approach for theme-aware 3D-to-3D generation and expanding the scale of existing 3D models. Code and video are at \url{https://3dthemestation.github.io/}.}
  \label{fig:teaser}
\end{teaserfigure}

\maketitle

\section{Introduction}
In applications such as virtual reality or video games, we often need to create a large number of 3D models that are thematically consistent with each other while being different. For example, we may need to create an entire 3D gallery of buildings to form an ancient town or monsters to form an ecosystem in a virtual world.
While it is easy for a highly trained craftsman to create one or a few coherent 3D models, it can be challenging and time-consuming 
to create a large 3D gallery.
We consider if we can automate this labor-intensive process, and
whether a generative system can produce 
many
unique 3D models that are different from each other while sharing a consistent style.

Recently, diffusion models~\cite{ho2020denoising} have revolutionized the 3D content creation task by significantly lowering the amount of manual work. This allows even beginners to create 3D assets from text prompts (\ie, text-to-3D) or reference images (\ie, image-to-3D) with minimal effort.  Early works~\cite{poole2022_dreamfusion} focus on using well-trained image diffusion models to generate 3D assets from a text prompt with score distillation sampling (SDS). Subsequent works~\cite{tang2023_makeit3d,melaskyriazi2023_realfusion} extend this approach to enable 3D creation from a single image. 
While these methods have shown impressive performances, they still suffer from the 3D ambiguity and inconsistency problem due to the limited 3D information from the input modality.

To address these limitations, in this work, we propose to leverage 3D exemplars as input to guide the 3D generation process. Given one or a few exemplar 3D models as input (\reffig{teaser} left), we present~\textbf{\textit{ThemeStation}}, a novel approach for the theme-aware 3D-to-3D generation task, which aims to generate a diverse range of unique 3D models that are theme-consistent (\ie, semantically and stylistically the same) with the input exemplars while being different from each other.
Compared to text prompts and images, 3D exemplars offer a richer source of information with respect to both geometry and appearance, reducing ambiguity in 3D modeling. This, in turn, makes it possible to create higher-quality 3D models.
ThemeStation enables the automatic synthesis of, for example, a group of buildings/characters with a shared theme (\reffig{teaser} right). 
It aims to satisfy two goals in the 3D generation process: \textbf{unity} and \textbf{diversity}. For unity,  we expect the generated models to align with the theme of the given exemplars. For diversity, we aim for the generated models to exhibit a high degree of variations.

However, we note that simply training a generative model on a few 3D exemplars~\cite{wu2022learning, wu2023sin3dm} leads to only limited variation, primarily restricted to resizing the input models (to different scales and aspect ratios) or repeating them randomly (\reffig{comparison_3d_to_3d}), without introducing significant modifications to the appearance of the generated models. To address this problem, we design a two-stage generative scheme to mimic the manual 3D modeling workflow of first drawing a concept art and then using a progressive 3D modeling process~\cite{StagesCreate3DModel2022, 3DModeling1012022}. 
In the first stage, we fine-tune a well-trained image diffusion model~\cite{rombach2022high} on rendered images of the given 3D exemplars to produce diverse concept images.
Unlike previous fine-tuning techniques~\cite{ruiz2023dreambooth,gal2022textual} that are subject-driven, our goal is to 
personalize the pre-trained diffusion model with a specific theme to synthesize images with novel subjects. 
In the second stage, we convert the synthesized concept images into 3D models. Our setting differs from image-to-3D tasks in that (1) we only regard the concept images as intermediate outputs to provide rough guidance on the overall structure and appearance of the generated 3D models and (2) we take the input 3D exemplars as auxiliary guidance to provide additional geometry and multi-view appearance information.
To leverage both the synthesized concept images and input 3D exemplars (also referred to as the reference models in this paper), 
we propose reference-informed \textbf{dual score distillation (DSD)} to guide the 3D modeling process using two diffusion models: one (\textbf{Concept Prior}) for enforcing content fidelity in concept image reconstruction, similar to~\cite{raj2023dreambooth3d}, and the other (\textbf{Reference Prior}) for reconstructing multi-view consistent fine details from the exemplars.
Instead of naively combining the two losses, which may lead to severe loss conflict, we apply
the two priors based on the noise levels (denoising timesteps). 
While the concept prior is applied to high noise levels for guiding the global layout, the reference prior is applied to low noise levels for guiding low-level variations.

To evaluate our approach,
we have collected a benchmark that contains stylized 3D models with varying complexity.
As shown in Fig.~\ref{fig:teaser}, \textit{ThemeStation} can produce a creative gallery of 3D assets conforming to the theme of the input exemplars. Extensive experiments and a user study show that \textit{ThemeStation} can generate compelling and diverse 3D models with finer details, even with just a single input exemplar. 
\textit{ThemeStation} also enables various applications, such as controllable 3D-to-3D generation, showing immense potential for generating creative 3D content and expanding the scale of existing 3D models.
Our main contributions can be summarized as:
\begin{itemize}
    \item We propose \textit{ThemeStation}, a two-stage framework for theme-aware 3D-to-3D generation,
    which aims at generating novel 3D assets with unity and diversity given just one or a few 3D exemplars.
    \item We make a first attempt to tackle the challenging problem of extending diffusion priors for 3D-to-3D content generation.
    \item We introduce dual score distillation (DSD) to enable the joint usage of two conflicted diffusion priors for 3D-to-3D generation by applying the reference prior and concept prior at different noise levels.
\end{itemize}

\section{Related Work}
\subsection{3D Generative Models}
Remarkable advancements have been made to generative adversarial networks (GANs) and diffusion models for image synthesis~\cite{rombach2022high,saharia2022photorealistic,brock2018large,karras2019style}. Many researchers have explored how to apply these methods to generate 3D geometries using different representations, such as point clouds~\cite{nichol2022_pointe,zhou20213d}, meshes~\cite{nash2020polygen,pavllo2021learning} and neural fields~\cite{chan2022eg3d,niemeyer2021giraffe,erkoç2023hyperdiffusion}. Recent works can further generate 3D textured shapes~\cite{jun2023shap,wang2022_rodin,chen2023single,gupta20233dgen,hong20243dtopia,tang2024lgm}. These methods require a large 3D  dataset for training, which limits their performance on in-the-wild generation.

\subsection{Diffusion Priors for 3D Generation} 
Dreamfusion~\cite{poole2022_dreamfusion}  proposed to distill the score of image distribution from a pre-trained text-to-image (T2I) diffusion model and show promising results in text-to-3D generation. Subsequent works enhance the score distillation scheme~\cite{poole2022_dreamfusion} and achieve higher generative quality for text-to-3D generation~\cite{chen2023fantasia3d,lin2023_magic3d,metzer2023latent}. Some recent works also apply the diffusion priors to image-to-3D generation~\cite{melaskyriazi2023_realfusion,tang2023_makeit3d,sun2023dreamcraft3d,chen2024comboverse,tang2023dreamgaussian}.
To enhance multi-view consistency of the generated 3D content, some researchers seek to fine-tune the pre-trained image diffusion models with multi-view datasets for consistent multi-view image generation~\cite{shi2023_MVDream,long2023wonder3d,liu2023syncdreamer,liu2023one2345++}. 
Although diffusion priors have shown great potential for 3D content generation from text or image inputs, their applicability to 3D customization based on 3D exemplars is still an open and challenging problem.

\subsection{Exemplar-Based Generation}
The exemplar-based 2D image generation task has been widely explored~\cite{gal2022image,ruiz2023dreambooth,avrahami2023break}. 
Recently, 
DreamBooth3D~\cite{raj2023dreambooth3d} fine-tunes pre-trained diffusion models with only a few images to achieve subject-driven text-to-3D generation but still suffers inconsistency due to the lack of  3D information from the input images. Another line of work takes 3D exemplars as input to generate 3D variations. 
For example, assembly-based methods~\cite{zheng2013smart,chaudhuri2011probabilistic,kim2013learning,schor2019componet,xu2012fit} focus on
retrieving compatible parts from a collection of 3D examples and organizing them into a target shape. 
Some methods extend the idea of 2D SinGAN~\cite{shaham2019singan} to train a 3D generative model~\cite{wu2022learning,wu2023sin3dm} with a single 3D exemplar. 
Some methods~\cite{li2023patch} lift classic 2D patch-based frameworks
to 3D generation without the need for offline training. 
While these methods support 3D variations of sizes and aspect ratios, they do not understand and preserve the semantics of the 3D exemplars. 
As a result, their results are primarily restricted to resizing, repeating, or reorganizing the input exemplars in some way (\reffig{comparison_3d_to_3d}), which is different from our setting that aims \red{to produce theme-consistent} 3D variations.

\begin{figure*}[t]
    \centering
    \includegraphics[width=\linewidth]{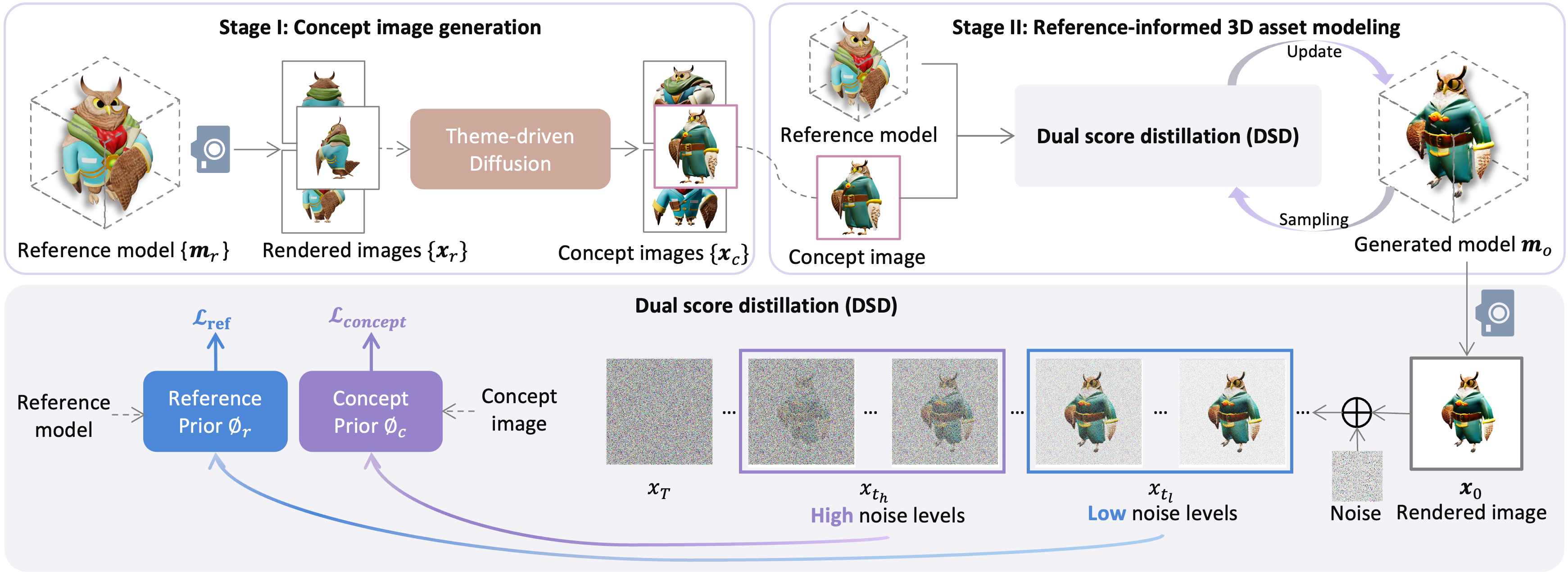}
    \caption{Overview of \textit{ThemeStation}. Given just one or a few reference models, our approach can generate theme-consistent 3D \red{models} in two stages. In the first stage, we fine-tune a pre-trained text-to-image (T2I) diffusion model to form a customized theme-driven diffusion model that produces various concept images. In the second stage, we conduct reference-informed 3D asset modeling by progressively optimizing a rough initial model \red{(omitted in this figure for brevity), which is obtained using an off-the-shelf image-to-3D method given the concept image,} into a final 3D asset. We use a novel dual score distillation (DSD) loss for optimization, which applies concept prior and reference prior at different noise levels (denoising timesteps).
    }
    \label{fig:overview}
\end{figure*}

\section{Approach}
\label{sec:approach}
 Our framework is designed to follow the real-world workflow of 3D modeling by introducing a concept art design step before the 3D modeling process. As illustrated in \reffig{overview}, we first customize a pre-trained text-to-image (T2I) diffusion model to produce a series of concept images that share a consistent theme as the input exemplars, mimicking the concept art designing process in practice (Sec.~\ref{sec:cig}). We then utilize an optimization-based method to lift \red{each concept image to} a final 3D model, following the practical modeling workflow of pushing a base primitive into a well-crafted 3D model (Sec.~\ref{sec:rim}). To this end, we present novel dual score distillation (DSD) that leverages the priors of both the concept images and the exemplars in the optimization process (Sec.~\ref{sec:DSD}).

\subsection{Theme-Driven Concept Image Generation}
\label{sec:cig}
Concept image design is a visual tool to convey the idea and preview the final 3D model. It is usually the first step in the 3D modeling workflow and serves as a bridge between the designer and the modeler~\cite{StagesCreate3DModel2022,3DModeling1012022}. 
Following this practice, in this stage, our goal is to generate a variety of concept images 
$\{\boldsymbol{x}_c\}$  of a specific theme based on the input exemplars $\{\boldsymbol{m}_r\}$,
% $\{\boldsymbol{m_0},\dots,\boldsymbol{m_N}\}$
% , where $M>0$ and $N>0$ are the number of synthesized reference images and input examples, respectively, 
as shown in~\reffig{overview} top. While there are some existing works on subject-driven image generation~\cite{ruiz2023dreambooth, gal2022textual}, which fine-tune a pre-trained T2I diffusion model~\cite{rombach2022high} to generate novel contexts for a specific (exactly the same) subject, they are not aligned with our theme-driven setting. Our goal is to generate a diverse set of subjects that exhibit thematic consistency but display content variations relative to the exemplars. Thus, instead of stimulating the subject retention capability of the pre-trained diffusion model through overfitting the inputs, we seek to preserve its imaginative capability while \red{preserving the theme of the} input exemplars.

We observe that the diffusion model, fine-tuned with fewer iterations \red{on the rendered images $\{\boldsymbol{x}_r\}$ of the input exemplars $\{\boldsymbol{m}_r\}$}, \red{is already able to learn the theme of the exemplars}. 
Hence, it is able to generate novel subjects that are \red{thematically} in line with the input exemplars.
To further disentangle the \red{theme (semantics and style) and the} content (subject) of the exemplars, we explicitly indicate the learning of \red{the theme} using a shared text prompt across all exemplars,~\eg~``a 3D model of an owl, in the style of [V]'', \red{during the fine-tuning process}.

\subsection{Reference-Informed 3D Asset Modeling}
\label{sec:rim}
Given one synthesized concept image  ${\boldsymbol{x}_c}$ and the input exemplars $\{\boldsymbol{m}_r\}$, we conduct reference-informed 3D asset modeling in the second stage.
\red{Similar to the workflow of practical 3D modeling that starts with a base primitive}, we begin with a rough initial 3D model $\boldsymbol{m}_{init}$, generated using off-the-shelf image-to-3D techniques~\cite{liu2023zero,liu2023syncdreamer,long2023wonder3d} given the concept image ${\boldsymbol{x}_c}$, \red{to accelerate our 3D asset modeling process}.  
As the synthesized concept image, along with the initial 3D model, may have inconsistent spatial structures and unsatisfactory artifacts, we do not enforce \red{our final generated} model to be strictly aligned with the concept image.
\red{We then take} the concept image and the initial model as intermediate outputs and meticulously develop the initial model into \red{the final generated 3D model} $\boldsymbol{m}_{o}$.
Different from previous optimization-based methods that perform score distillation sampling using a single diffusion model~\cite{poole2022_dreamfusion,wang2023_prolificdreamer}, we propose a dual score distillation (DSD) loss to leverage two diffusion priors as guidance simultaneously. Here, one diffusion model, denoted as $\phi_{c}$, functions as the basic concept (\textbf{concept prior}), providing diffusion priors from the concept image ${\boldsymbol{x}_c}$ to ensure concept reconstruction, while the other, denoted as $\phi_{r}$, operates as an advisory reference (\textbf{reference prior}), generating diffusion priors pertinent to the input reference models $\{\boldsymbol{m}_r\}$ to assist with restoring subtle features and alleviating multi-view inconsistency.
We further present a clear design of our DSD loss in~\refsec{DSD}.

\begin{figure}[t]
    \centering
    \includegraphics[width=\linewidth]{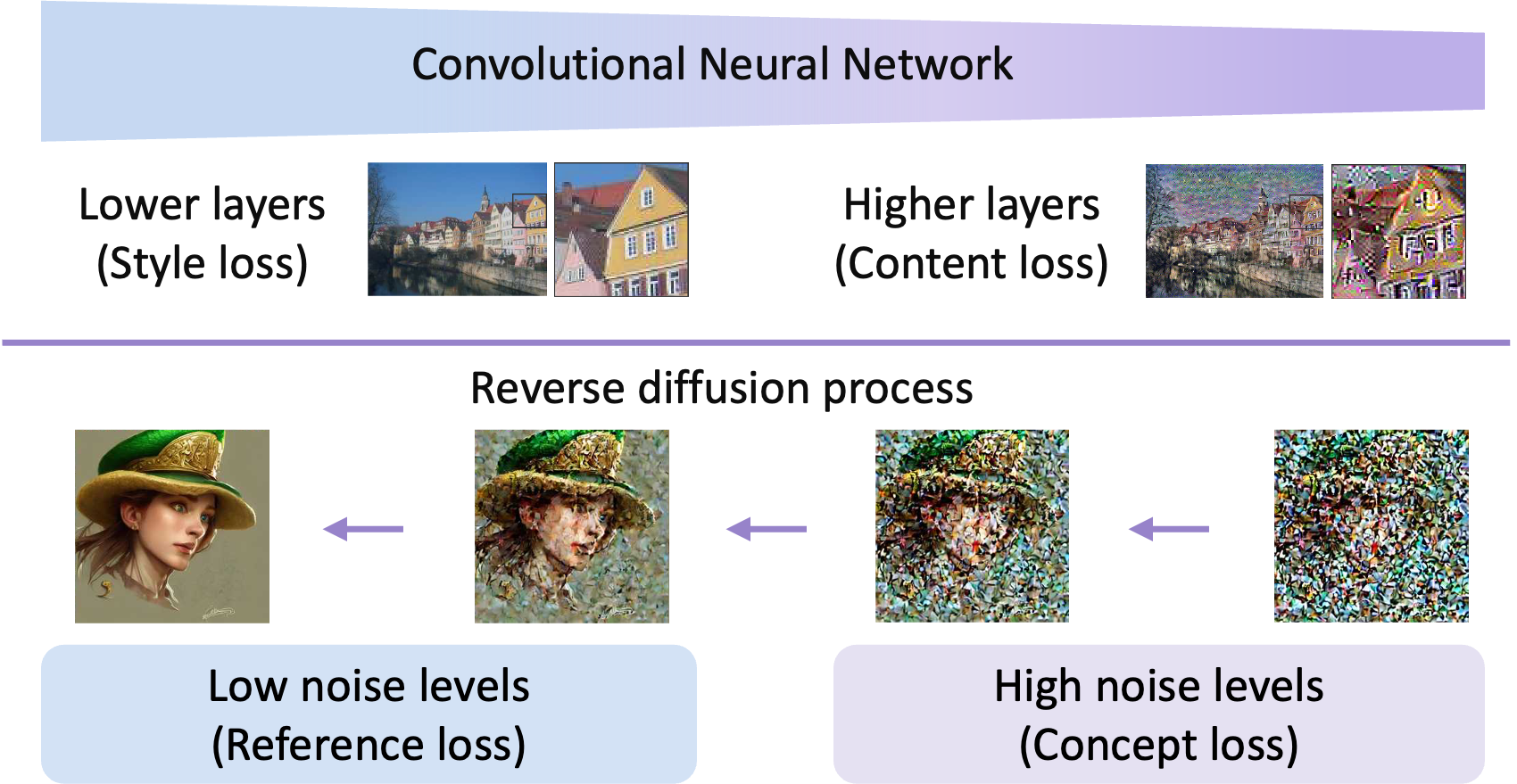}
    \caption{Comparison of the key ideas between image style transfer (top) and our dual score distillation (bottom). Images are from Gatys~\etal~\shortcite{gatys2016image} (top) and Dibia~\shortcite{DenoisingSteps2022} (bottom).}
    \label{fig:style_transfer}
\end{figure}

\subsection{Dual Score Distillation}
\label{sec:DSD}
In this subsection, we elaborate on the critical component of our approach, dual score distillation (DSD) for theme-aware 3D-to-3D generation. DSD  combines the best of both priors, concept prior and reference prior, to guide the generation process. Both priors are derived through fine-tuning a pre-trained T2I diffusion model. Next, we discuss the preliminaries and show the steps of learning the two priors and the design of DSD loss.

% \noindent\textbf{Preliminaries.}
\subsubsection{Preliminaries}
DreamFusion achieves text-to-3D generation by optimizing a 3D representation with parameter $\theta$ so that the \red{randomly} rendered images $\boldsymbol{x}=g(\theta)$ under different camera poses look like 2D samples of a pre-trained T2I diffusion model for a given text prompt $y$. Here, $g$ is a NeRF-like rendering engine. The T2I diffusion model $\phi$ works by predicting the sampled noise ${\epsilon}_\phi\left(\boldsymbol{x}_t ; y, t\right)$ of a rendered view $\boldsymbol{x}_t$ at noise level $t$ for a given text prompt $y$. To move all rendered images to a higher density region under the text-conditioned diffusion prior, score distillation sampling (SDS) estimates the gradient for updating $\theta$ as:
\begin{equation}
\label{eq:sds}
  \nabla_\theta \mathcal{L}_{\mathrm{SDS}}(\phi, x)=\mathbb{E}_{t, {\epsilon}}\left[\omega(t)\left({\epsilon}_\phi\left(\boldsymbol{x}_t ; y, t\right)-{\epsilon}\right) \frac{\partial \boldsymbol{x}}{\partial \theta}\right] ,
\end{equation}
where $\omega(t)$ is a weighting function. 

Following SDS, variational score distillation (VSD)~\cite{wang2023_prolificdreamer} further improves generation diversity and quality, which regards the text-conditioned 3D representation as a random variable rather than a single data point in SDS. The gradient is computed as: 
\begin{equation}
\label{eq:vsd}
    \nabla_\theta \mathcal{L}_{\mathrm{VSD}}=\mathbb{E}_{t, {\epsilon}}\left[\omega(t)\left({\epsilon}_\phi\left(\boldsymbol{x}_t ; y, t\right)-{\epsilon}_{\mathrm{lora}}\left(\boldsymbol{x}_t ; y, t, c\right)\right) \frac{\partial \boldsymbol{x}}{\partial \theta}\right] ,
\end{equation}
where $c$ is the camera parameter, and ${\epsilon}_{\mathrm{lora}}$ computes the score of noisy rendered images by a low-rank adaption (LoRA) \cite{hu2021lora} of the pre-trained T2I diffusion model.
Despite the promising quality, both VSD and SDS mainly work on distilling the unitary prior from a single diffusion model and may collapse when encountering mixed priors from conflicted diffusion models.

% \noindent\textbf{Learning of concept prior.}
\subsubsection{Learning of concept prior}
To learn concept prior, we leverage not only the concept image itself but also the 3D consistent information in its initial 3D model $\boldsymbol{m}_{init}$. We observe that the initial model suffers blurry texture and over-smoothed geometry, which is insufficient to provide a high-quality concept prior. 
Thus, we augment the initial rendered views $\{\boldsymbol{x}_{init}\}$ of $\boldsymbol{m}_{init}$ into augmented views $\{\hat{\boldsymbol{x}_{init}}\}$,~\ie~$\{\hat{\boldsymbol{x}_{init}}\}=a(\{\boldsymbol{x}_{init}\})$, where $a(\cdot)$ is the image-to-image translation operation, similar to~\cite{raj2023dreambooth3d}. These augmented views serve as pseudo-multi-view images of the conceptual subject, providing additional 3D information for further 3D modeling.  
Finally, the diffusion model ${\phi_{c}}$ with concept prior is derived by fine-tuning a T2I diffusion model given $\{x_c, \{\hat{\boldsymbol{x}_{init}}\},y\}$, where $y$ is the text prompt with a special identifier,~\eg~``a 3D model of [V] owl''.

% \noindent\textbf{Learning of reference prior.}
\subsubsection{Learning of reference prior}
To learn reference prior, we leverage both the color images ${\{\boldsymbol{x}_r}\}$ and the normal maps $\{\boldsymbol{n}_r\}$ rendered from the reference models $\{\boldsymbol{m}_r\}$ under random viewpoints. While the rendered color images mainly provide 3D consistent priors on textures, the rendered normal maps focus on encoding detailed geometric information. The joint usage of these two kinds of renderings helps to build up a more comprehensive reference prior for introducing 3D consistent details during optimization.   To disentangle the learning of image prior and normal prior, we also incorporate different text prompts, $y_x$ and $y_n$, for color images,~\eg~``a 3D model of an owl, in the style of [V]'', and normal maps,~\eg~``a 3D model of an owl, in the style of [V], normal map'', respectively. Finally, the diffusion model $\phi_r$ with reference prior is derived by fine-tuning a pre-trained T2I diffusion model given $\{\{\boldsymbol{x}_r\},y_x,\{\boldsymbol{n}_r\},y_n\}$. Although we convert the 3D reference models into 2D space, their 3D information has still been implicitly reserved across the consistent multi-view rendered color images and normal maps. \red{Besides, as the pre-trained T2I diffusion models have been shown to possess rich 2D and 3D priors about the visual world~\cite{liu2023zero}, we can also inherit these priors to enhance our modeling quality by projecting the 3D inputs into 2D space.}

% \noindent\textbf{How does dual score distillation work?}
\subsubsection{How does dual score distillation work?}
A straightforward aggregation of these two priors is performing the vanilla score distillation sampling twice indiscriminately for both diffusion models $\phi_c$ and $\phi_r$ and summing up the losses. However, this naive stack of two priors leads to loss conflicts during optimization and generates distorted results ((b) of \reffig{ablation_study}). To resolve this, we introduce a dual score distillation (DSD) loss, which applies the two diffusion priors at different noise levels (denoising timesteps) during the reverse diffusion process. 

This method is based on our observation that there is a coarse-to-fine timestep-based dynamic during the reverse diffusion process. High noise levels,~\ie~the early denoising timesteps $t_{h}$, control the global layout and rough color distribution of the image being denoised. As the reverse diffusion gradually goes into low noise levels,~\ie~the late denoising timesteps $t_{l}$, high-frequency details are generated. This intriguing timestep-based dynamic process of T2I diffusion models is incredibly in line with the functionalities of our concept prior and reference priors. Inspired by image style transfer~\cite{gatys2016image} that leverages different layers of a pre-trained neural network to control different levels of image content, as shown in~\reffig{style_transfer}, we apply the concept prior $\phi_c$ at high noise levels $t_h$ to enforce the concept fidelity by adjusting the layout and color holistically, and apply the reference prior $\phi_r$ at low noise levels $t_l$ to recover the finer elements in detail. 

Based on \refeq{vsd}, the gradient for updating the 3D representation $\theta$ of the model being optimized given the concept prior is:
\begin{equation}
\label{eq:DSD_concept}
\nabla_\theta \mathcal{L}_{\mathrm{concept}}(\phi_c,t_h)=\mathbb{E}_{t_h, {\epsilon}}\left[\omega\left({\epsilon}_{\phi_c}\left(\boldsymbol{x}_{t_h} ; y, t_h\right)-{\epsilon}_{\mathrm{lora}}\right) \frac{\partial \boldsymbol{x}}{\partial \theta}\right] ,
\end{equation}
where $\omega$ is a weighting function, ${\epsilon}_{\phi_c}\left(\boldsymbol{x}_{t_h} ; y, t_h\right)$ is the sampled noise of the rendered color image $\boldsymbol{x}_{t_h}$ at high noise level $t_h$ conditioned on prompt $y$, and $\epsilon_{lora}$ is the score of noisy rendered images parameterized by a LoRA of a pre-trained diffusion model. For reference prior, we apply it on both rendered color images and normal maps to jointly recover the detailed texture and geometry with the learned image prior and normal prior from the reference models. 
The gradient given the reference prior is:
\begin{equation}
\label{eq:DSD_ref}
\begin{aligned}
\nabla_\theta \mathcal{L}_{\mathrm{ref}}(\phi_r,t_l)&=
\mathbb{E}_{t_l, {\epsilon}}\left[\omega\left({\epsilon}_{\phi_r}\left(\boldsymbol{x}_{t_l} ; y_x, t_l\right)-{\epsilon}_{\mathrm{lora}}\right) \frac{\partial \boldsymbol{x}}{\partial \theta}\right]
\\& + 
\mathbb{E}_{t_l, {\epsilon}}\left[\omega\left({\epsilon}_{\phi_r}\left(\boldsymbol{n}_{t_l} ; y_n, t_l\right)-{\epsilon}_{\mathrm{lora}}\right) \frac{\partial \boldsymbol{x}}{\partial \theta}\right] ,
\end{aligned}
\end{equation}
where $\boldsymbol{x}_{t_l}$ and $\boldsymbol{n}_{t_l}$ are the rendered color image and normal map at low noise level $t_l$, and $y_x$ and $y_n$ are their corresponding text prompts. Finally, the gradient of our DSD loss is:
\begin{equation}
\label{eq:DSD}
    \nabla_\theta \mathcal{L}_{\mathrm{DSD}}=\alpha \nabla_\theta \mathcal{L}_{\mathrm{concept}}(\phi_{c},t_h)
    +\beta \nabla_\theta \mathcal{L}_{\mathrm{ref}}(\phi_{r},t_l),
\end{equation}
where $\alpha$ and $\beta$ are weights to balance the strength of two guidance.

\section{Experiments}
We show the generated 3D models based on a few 3D exemplars \revision{in~\reffig{group_results}}. We can see that our approach can generate various novel 3D assets that share consistent themes with the input exemplars. These generated 3D assets exhibit finer texture and elaborate geometry, ready for real-world usage (\reffig{teaser}). Our approach can even work with only one exemplar, as shown \revision{in~\reffig{single_results}}.
For the rest of this section, we first conduct experiments and a user study to compare our results with those generated by the state-of-the-art methods. We also conduct experiments to analyze the effectiveness of several important design choices of our approach. We show implementation details in supplementary materials.

\subsection{Comparisons with State-of-the-Art Methods}
\subsubsection{Benchmark} 
\begin{figure*}[!t]
    \centering
    \includegraphics[width=\linewidth]{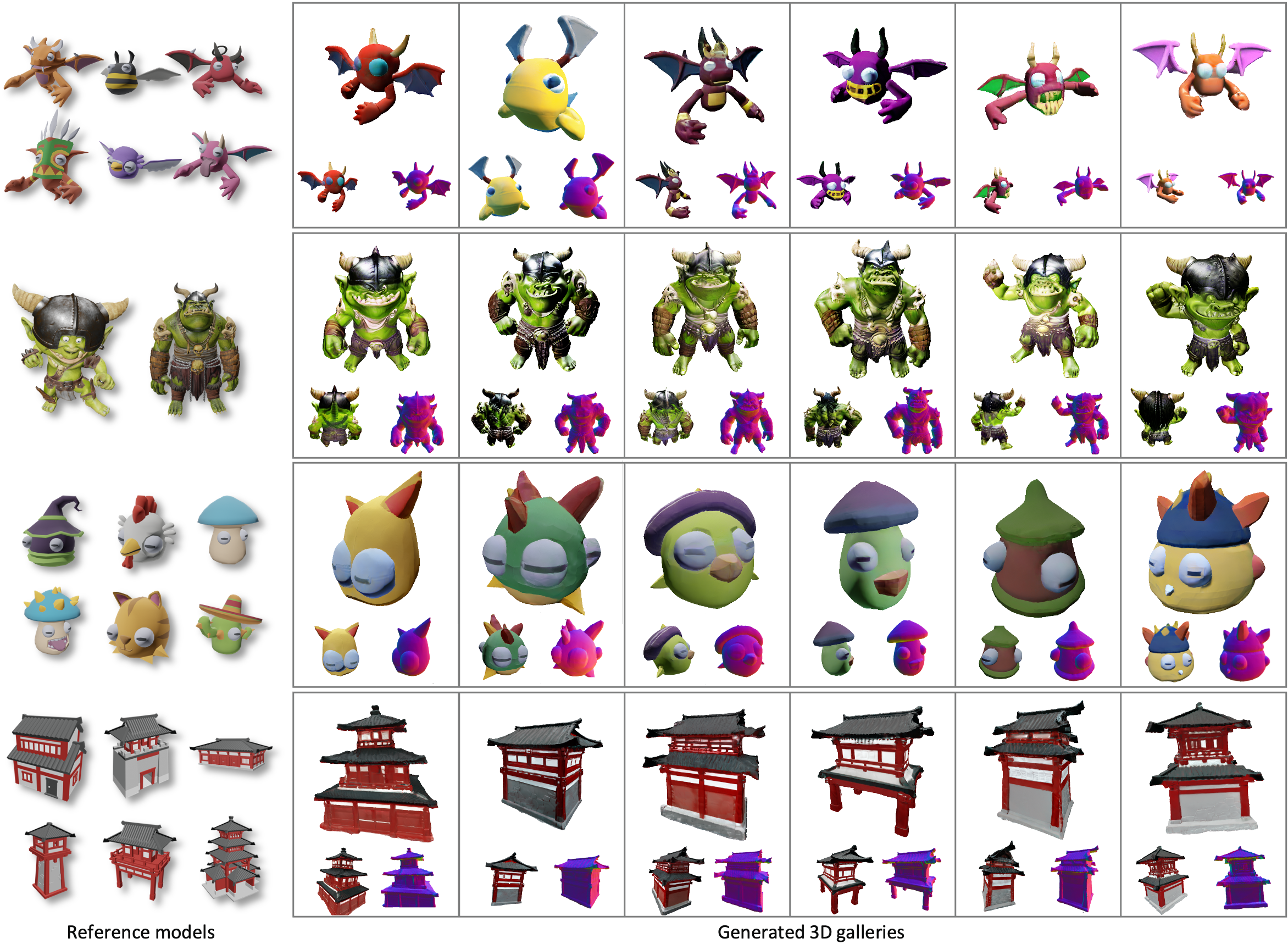}
    \caption{Visual results of \textit{ThemeStation}, which  generates 3D models from a few 3D exemplars. For each case, we show the reference models on the left and six generated models on the right. For each generated model, we show a primary view (top) with its normal map (bottom right) and a secondary view (bottom left).}
    \label{fig:group_results}
\end{figure*}

\begin{figure*}[!t]
    \centering
    \includegraphics[width=\linewidth]{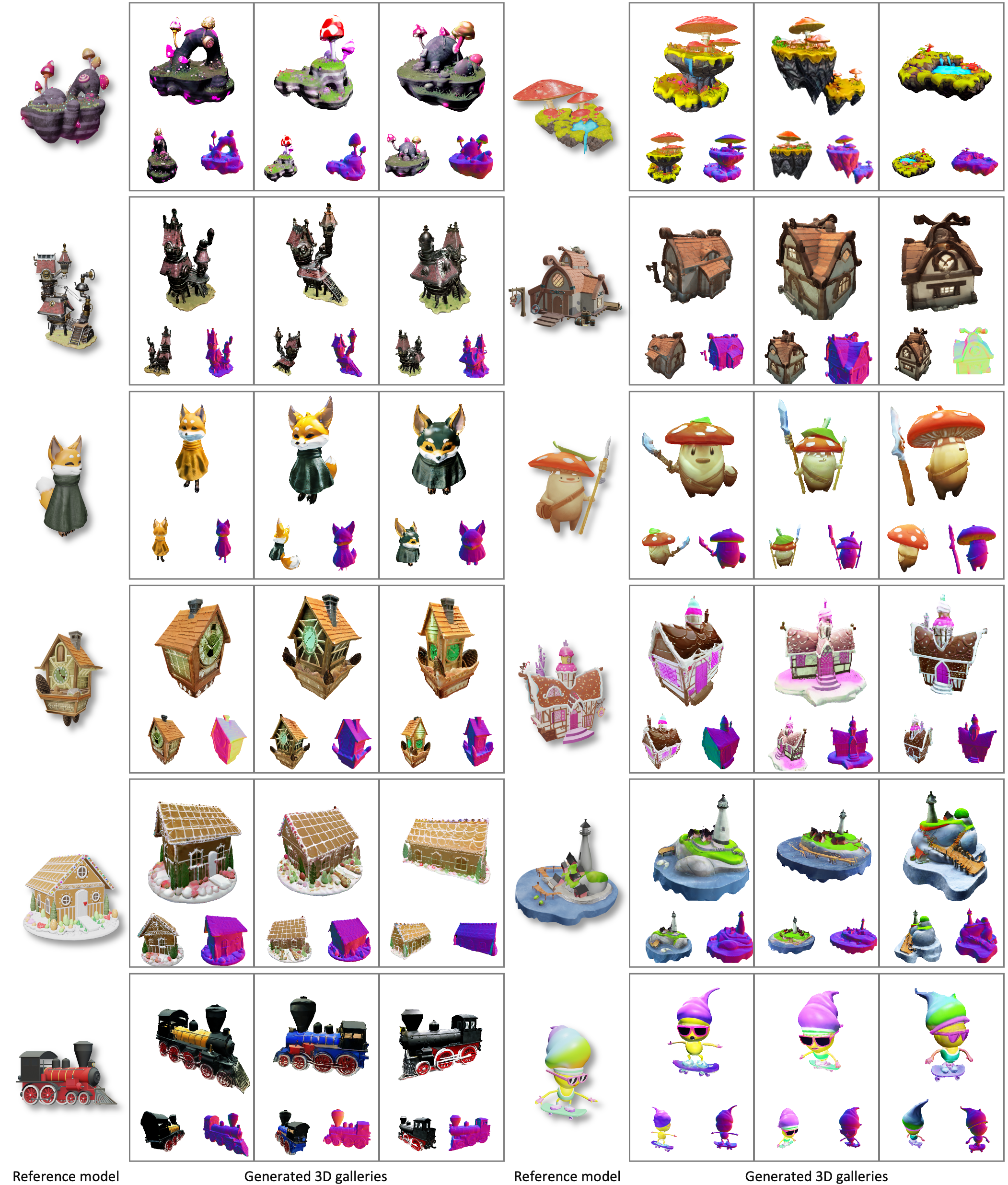}
    \caption{Visual results of \textit{ThemeStation}, which  generates 3D models from only one 3D exemplar. For each case, we show the reference model (left) and three generated models on the right. For each generated model, we show a primary view (top) with its normal map (bottom right) and a secondary view (bottom left).}
    \label{fig:single_results}
\end{figure*}

% 40 (single) + 26(group)
We have collected a dataset of 66 reference models covering a broad range of themes. These 3D models comprise three main categories, including $15$ dioramas, $25$ individual objects, and $26$ characters, such as small islands, buildings, and characters, as shown in~\reffig{group_results}-\ref{fig:single_results}. Models in this dataset are exported from the built-in 3D library of \textit{Microsoft 3D Viewer} or downloaded from \textit{Sketchfab}\footnote{\url{https://sketchfab.com/}}. The text prompts for each 3D model are automatically generated by feeding the model's subject name,~\ie~file name in most cases, into the pre-defined patterns presented in~\refsec{approach}.

\subsubsection{Methods Compared}
\red{To the best of our knowledge, ours is the first work focusing on theme-aware 3D-to-3D generation with diffusion priors.}
\revision{As no existing methods can simultaneously take both images and 3D models as inputs,} we compare our method with seven baseline methods from two aspects. On the one hand, we compare with five \textbf{image-to-3D} methods\revision{, including 
multi-view-based,~\ie~\textit{Wonder3D}~\cite{long2023wonder3d}, ~\textit{SyncDreamer (SyncD.)}~\cite{liu2023syncdreamer}, 
feed-forward,~\ie~\textit{LRM}~\cite{hong2023lrm}, ~\textit{Shape-E}~\cite{jun2023shap}, 
and optimization-based,~\ie~\textit{Magic123}~\cite{qian2023_magic123},} to evaluate our second stage that lifts a concept image to a 3D model. 
Due to the unavailable code of LRM, we use its open-source reproduction ~\textit{OpenLRM}~\cite{openlrm}.
On the other hand, we also compare with two \textbf{3D variation} methods: ~\textit{Sin3DM}~\cite{wu2023sin3dm}  and ~\textit{Sin3DGen}~\cite{li2023patch}, to evaluate the \red{overall 3D-to-3D performance} of our method.

\begin{table}
\setlength\tabcolsep{3pt}
  \caption{Quantitative comparison with  image-to-3D methods.}
  \label{tab:comparison_image_to_3d}
  \footnotesize
  \resizebox{\linewidth}{!}{
  \begin{tabular}{lcccccc}
    \toprule
     & Wonder3D & OpenLRM & SyncD. & Shap-E & Magic123 & Ours
    \\
    \midrule
    CLIP $\uparrow$ & 0.777 & 0.840 & 0.803 & 0.761 & 0.868 & \textbf{0.890}
    \\
    \midrule
    Contextual $\downarrow$ & 3.206 & 4.137 & 4.189 & 3.399 & 3.345 & \textbf{3.168} 
    \\
  \bottomrule
\end{tabular}}
\end{table}

\begin{table}
  \caption{Quantitative comparison with  3D variation methods.}
  \label{tab:comparison_3d_to_3d}
  \footnotesize
  \begin{tabular}{lcccc}
    \toprule
     & Sin3DM & Sin3DGen & Ours \\
    \midrule
    Visual Diversity $\uparrow$ & 0.180 & 0.201 & \textbf{0.315}  \\
    \midrule
    Geometry Diversity $\uparrow$ & 0.344 & \textbf{0.634} & 0.465 \\
    \midrule
    \revision{Visual Quality $\uparrow$} & 5.221 & 5.127 & \textbf{5.848}  \\
    \midrule
    \revision{Geometry Quality $\uparrow$} & \textbf{5.638} & 5.607 & 5.616  \\
  \bottomrule
\end{tabular}
\end{table}

\subsubsection{Quantitative Results.}

\begin{figure*}[!t]
    \centering
    \includegraphics[width=0.97\linewidth]{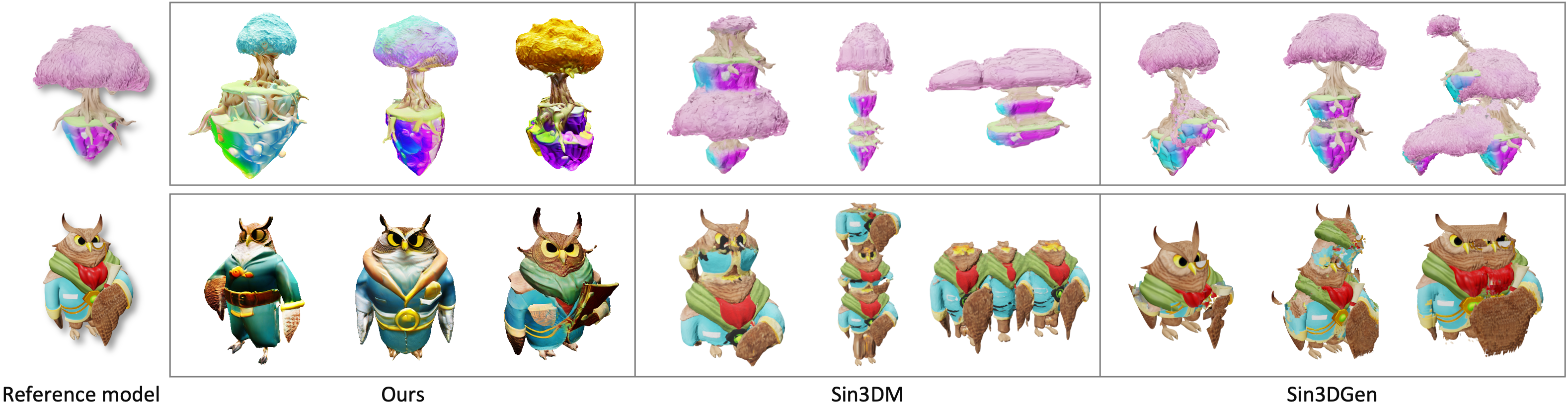}
    \caption{Qualitative comparisons with two 3D variation methods to evaluate the overall generative diversity and quality of our method. For each case, we show three generated 3D models.}
    \label{fig:comparison_3d_to_3d}
\end{figure*}

\textbf{For image-to-3D}, as our approach is not targeted to strictly reconstruct the input view, we focus on evaluating the semantic coherence between the input view and randomly rendered views of generated models. Thus, we adopt two metrics: 1) \textit{CLIP} score~\cite{radford2021learning} to measure the global semantic similarity, and 2) \textit{Contextual} distance~\cite{mechrez2018contextual} to estimate the semantic distance at the pixel level. Both metrics are commonly used in image-to-3D~\cite{tang2023_makeit3d,sun2023dreamcraft3d}. 
\textbf{For 3D-to-3D}, we use the pairwise IoU distance (1-IoU) among generated models and the average LPIPS score across different views to measure the \textit{Visual Diversity} and \textit{Geometry Diversity}, respectively. \revision{To measure the \textit{Visual Quality} and \textit{Geometry Quality}, we use the LAION\footnote{\url{https://laion.ai/blog/laion-aesthetics/}} aesthetics predictor to predict the visual and geometry aesthetics scores given the multi-view rendered images (visual) and normal maps (geometry).} The quantitative results in~\reftab{comparison_image_to_3d} and~\reftab{comparison_3d_to_3d} show that our approach surpasses the baselines in generative diversity, quality and multi-view semantic coherency. Sin3DGen generates variations at the patch level, achieving higher geometry diversity.  \revision{Sin3DM generates variations via a diffusion model trained with only one exemplar, achieving higher geometry quality. However, both methods tend to overfit the input and generate meaninglessly repeated or reorganized contents with lower visual diversity and quality (\reffig{comparison_3d_to_3d}).} In contrast, ours generates theme-consistent novel 3D assets with diverse and plausible variations in terms of both geometry and texture.

\begin{figure}[!t]
    \centering
    \includegraphics[width=\linewidth]{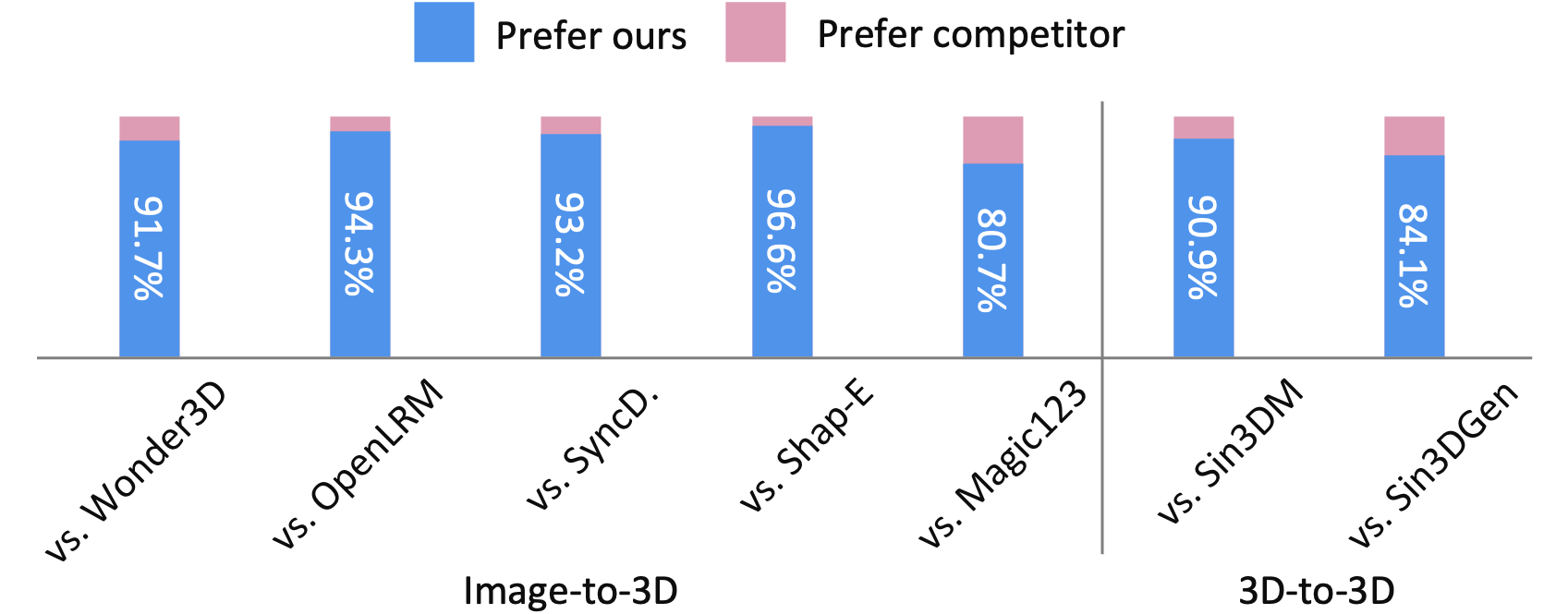}
    \caption{Results of the user study. We compare our method with seven baseline methods using 2AFC pairwise comparisons. All preferences are statistically significant ($p<0.05$, chi-squared test).}
    \label{fig:user_study}
\end{figure}

\subsubsection{User Study.}

The metrics used above mainly measure the input-output similarity and pixel/voxel-level diversity, which are not able to present the overall performance of different methods. We thus conduct a user study to estimate real-world user preferences.  
We invite 30 users publicly to complete a questionnaire for pairwise comparisons. We explain the detailed settings of this user study in supplementary materials.  
We can see from ~\reffig{user_study} that our approach significantly outperforms existing methods in both image-to-3D and 3D-to-3D tasks in terms of human preferences.

\begin{figure*}[!htb]
    \centering
    \includegraphics[width=0.96\linewidth]{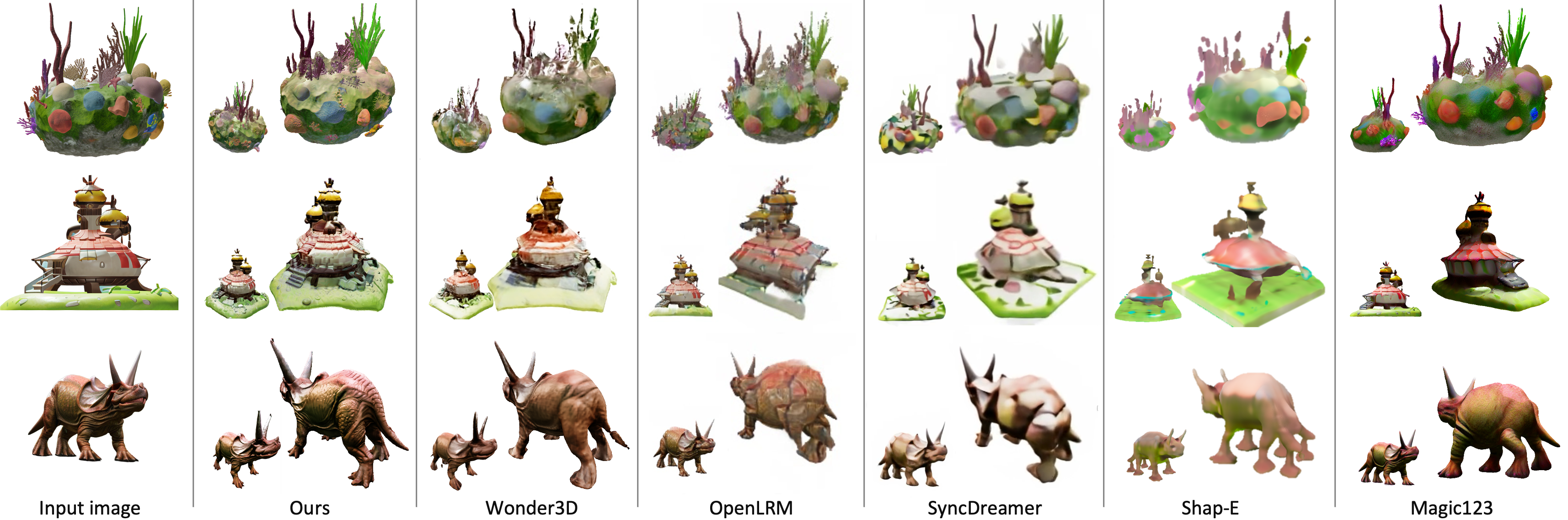}
    \caption{Qualitative comparisons with five image-to-3D methods to evaluate our second stage that lifts a concept image to a 3D model. We show the frontal view as primary for the first line and show the back view as primary for the last two lines.}
    \label{fig:comparison_image_to_3d}
\end{figure*}

\subsubsection{Qualitative Results.}
\label{sec:qualitative}

For image-to-3D comparison (\reffig{comparison_image_to_3d}), we can see that Shap-E, SyncDreamer, and OpenLRM suffer from lower quality with incomplete shape, blurry appearance, and multi-view inconsistency. Results of Wonder3D and Magic123 can generate 3D consistent models with higher quality. However, Wonder3D still generates vague texture and incomplete shape,~\eg~the severed tail of the triceratops, and Magic123 has problems with oversaturation and oversmooth.
All baseline methods lack delicate details, especially in novel views,~\eg~epidermal folds in the last line. In contrast, ours generates multi-view consistent 3D models with more details in geometry and texture.
For 3D-to-3D comparison (\reffig{comparison_3d_to_3d}), we can see that the baseline methods tend to randomly resize, repeat, or reorganize the input, which may produce weird results,~\eg~multi-head character and stump above treetop. Due to their theme-unaware 3D representation learned from just a few exemplars, it is hard for them to preserve or even understand the semantics of the input 3D exemplars. Instead, our approach combines priors from input 3D exemplars and pre-trained T2I diffusion models, yielding diverse semantically meaningful 3D variations that exhibit significant modifications on content while thematically aligning with the input exemplars.

\begin{figure*}[!t]
    \centering
    \includegraphics[width=0.97\linewidth]{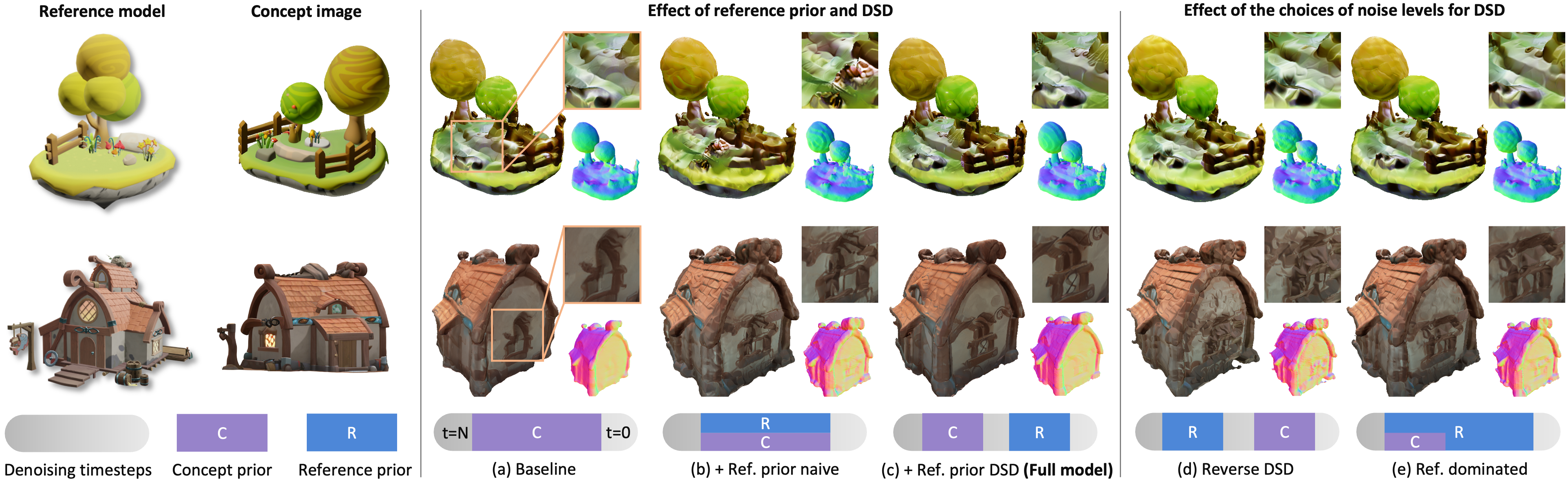}
    \vspace{-2mm}
    \caption{Ablation study on \revision{two types of effects: (1) reference prior and DSD loss (\refsec{ablation_ref_dsd}) and (2) the choice of noise levels for the DSD loss (\refsec{ablation_noise_level}).} (a) without the reference prior; (b) a naive combination of concept prior and reference prior; (c) using the proposed dual score distillation (DSD); (d) reversing the choice of noise levels for DSD; and (e) extending the reference prior to all noise levels for DSD. \revision{We show the back view for each case.}} 
    \vspace{-3mm}
    
    \label{fig:ablation_study}
\end{figure*}

\begin{table}
  \caption{\revision{Quantitative results of the ablation study.}}
  \label{tab:quantitative_ablation_study}
  \footnotesize
  \begin{tabular}{lccccc}
    \toprule
     & Baseline & +Ref. & +Ref. & Reverse & Ref. \\
     &  & naive & DSD & DSD & dominated \\
    \midrule
    CLIP $\uparrow$ & 0.877 & 0.876 & \textbf{0.890} & 0.863 & 0.874 \\
    \midrule
    Contextual $\downarrow$ & 3.182 & 3.177 & \textbf{3.168} & 3.186 & 3.179 \\
    \midrule
    Visual Quality $\uparrow$ & 5.639 & 5.726 & \textbf{5.848} & 5.578 & 5.701  \\
    \midrule
    Geometry Quality $\uparrow$ & 4.789 & 5.336 & \textbf{5.616} & 4.926 & 5.296 \\
  \bottomrule
\end{tabular}
\end{table}

\subsection{Ablation Study}

\subsubsection{Settings.}
To evaluate the effectiveness of our key design choices, we conduct ablation studies on five settings:  \textit{(a) Baseline}, which only uses concept prior across all noise levels,  \textit{(b) + Ref. prior naive}, which naively applies concept prior and reference prior across all noise levels,  \textit{(c) + Ref. prior DSD (full model)}, which applies concept prior at high noise levels and reference prior at low noise levels,  \textit{(d) Reverse DSD}, which reverses the choice of noise levels by applying concept prior at low noise levels and reference prior at high noise levels, and  \textit{(e) Ref. dominated}, which applies concept prior at high noise levels and reference prior across all noise levels. We measure the semantic coherence, visual quality and geometry quality as in image-to-3D and 3D-to-3D comparisons for the ablation study. As shown in~\reftab{quantitative_ablation_study}, our full model (+Ref. DSD) surpasses all baselines in terms of the four metrics mentioned above. We also show the qualitative comparison results in~\reffig{ablation_study}. Next, we further explore the scope and generality of the DSD loss.

\subsubsection{\revision{Effect of the reference prior and DSD loss.}}
\label{sec:ablation_ref_dsd}
As shown in \reffig{ablation_study}(a,c) and~\reftab{quantitative_ablation_study}, it is evident that the introduction of the reference prior and DSD significantly enhances the model quality \revision{in terms of semantic coherence, texture and geometry. From \reffig{ablation_study}(b), we can see that the naive combination of the reference prior and concept prior results in severe loss conflict and produces bumpy surface and blurry texture, which further} demonstrate the effectiveness of our DSD for alleviating loss conflicts. 

\subsubsection{\revision{Effect of the choices of noise levels for the DSD loss.}}
\label{sec:ablation_noise_level}
\revision{By comparing (c) with (d) in~\reffig{ablation_study}, we can see a significant performance degradation after reversing} the noise levels, which justifies our claim that the timestep-based dynamic process of T2I diffusion models is consistent with the functionalities of our concept prior and reference prior (\refsec{DSD}). Besides, \revision{by comparing (c) with (e) in~\reffig{ablation_study}, we can see that} extending the noise levels for the reference prior has no positive effect but leads to a worse result, indicating that the design of separating two priors at different noise levels can help reduce loss conflict.

\vspace{-1mm}
\section{Application}
\begin{figure*}[!t]
    \centering
    \includegraphics[width=\linewidth]{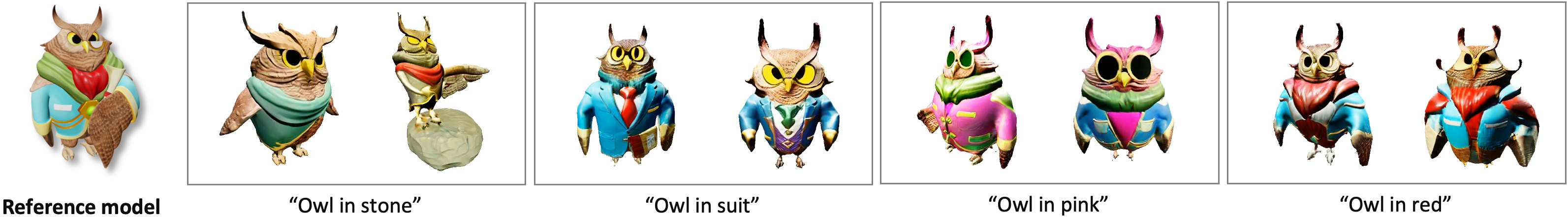}
    \caption{Application results of controllable 3D-to-3D generation. ~\textit{ThemeStation} allows users to specify a desired 3D variation via text prompt manipulation.}
    \label{fig:application}
\end{figure*}

As shown in~\reffig{application}, \textit{ThemeStation} supports the application of controllable 3D-to-3D generation, which allows users to control the concept image generation process via text prompt manipulation and obtain specific 3D variations. This sample application demonstrates the immense potential of \textit{ThemeStation} to be seamlessly combined with emerging controllable image generation techniques~\cite{wang2022pretraining, brooks2023instructpix2pix,hertz2022prompt} for more interesting 3D-to-3D applications.

\vspace{-1mm}
\section{Conclusion}
In this work, we proposed~\textit{ThemeStation}, a novel approach for the theme-aware 3D-to-3D generation task. Given just one or a few 3D exemplars, we aim to generate a gallery of unique theme-consistent 3D models.
% that share the same semantics and coherent styles with the input exemplars but different from each other. 
~\textit{ThemeStation} achieves this goal following a two-stage generative scheme that first draws a concept image as rough guidance and then converts it into a 3D model. Our 3D modeling process involves two priors, one from the input 3D exemplars (reference prior) and the other from the concept image (concept prior) generated in the first stage. A dual score distillation (DSD) loss function is proposed to disentangle these two priors and alleviate loss conflict. We have conducted a user study and extensive experiments to validate the effectiveness of our approach.

\begin{figure}[!t]
    \centering
    \includegraphics[width=0.97\linewidth]{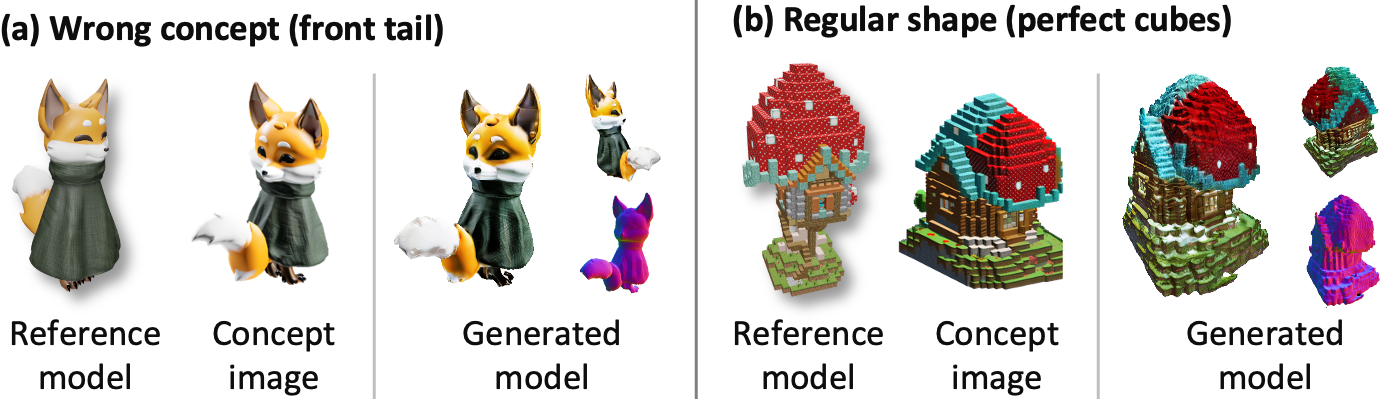}
    \vspace{-2mm}
    \caption{\revision{Failure cases. 
(a) Our approach may fail to fix huge concept errors when the concept image contains significant artifacts or mistakes,~\eg~the tail grows in front of the body. (b) Our approach may fail to generate perfect 3D models of regular shapes, such as a “Minecraft” building with cubic regularization, due to the lack of explicit geometry constraints.
}}
\vspace{-3mm}

    \label{fig:failure_case}
\end{figure}

% \revision{
% \noindent\textbf{Limitations and failure cases.} 
While~\textit{ThemeStation} produces high-quality 3D assets given just one or a few 3D exemplars and opens up a new venue for theme-aware 3D-to-3D generation, it still has several limitations for further improvement. First, similar to prior optimization-based 3D generation methods, it still takes hours for our current pipeline to optimize the initial model into a final 3D asset. We believe advanced diffusion models and neural rendering techniques in the future can help alleviate this problem. Besides, like two sides of a coin, as a two-stage pipeline, although~\textit{ThemeStation} can be easily adapted to emerging image-to-3D methods for obtaining a better initial model, it may also suffer from a bad initialization sometimes,~\eg~3D artifacts and floaters. Training a feed-forward theme-aware 3D-to-3D generation model is a potential solution, which can be an interesting future work. 
Failure cases are shown in~\reffig{failure_case}.

\vspace{-2mm}
\begin{acks}
This work is partially supported by the National Key R\&D Program of China (2022ZD0160201) and Shanghai Artificial Intelligence Laboratory.  This work is also in part supported by a GRF grant from the Research Grants Council of Hong Kong (Ref. No.: 11205620). 
\end{acks}

\clearpage

\bibliographystyle{ACM-Reference-Format}

\bibliography{sample-base}

\clearpage

\appendix

\section*{Supplementary Material}
  
\section{Implementation Details}
\label{sec:implementation}
\textbf{In the first stage}, we render 20 images for each reference model with a fixed elevation,~\ie~$0$ or $20$, and randomized azimuth. We fine-tune the pre-trained Stable Diffusion~\cite{rombach2022high} model for 200 iterations (a single exemplar) or 400 iterations (a few exemplars)  with a batch size of 8. We set the learning rate as $2\times10^{-6}$, the image size as $512\times512$, and the CFG weight at inference as $7.5$. We also take the camera pose of the rendered images as an additional condition during the model fine-tuning step to ensure the generated concept images have a correct viewpoint for accurate image-to-3D initialization.

\noindent\textbf{In the second stage}, we employ an off-the-shelf image-to-3D method~\cite{long2023wonder3d} to lift the synthesized concept image into an initial 3D model, represented as a neural implicit signed distance field (SDF). We use the concept image and 20 augmented views of the initial model for concept prior learning and use 30 normal maps, and 30 color images of the input 3D exemplars for reference prior learning. 
During optimization, we convert the SDF into DMTet~\cite{shen2021deep} at a 192 grid and 512 resolution to directly optimize the textured mesh at each optimization iteration. We render both the normal map and the color image, under randomized viewpoints, as guidance to compute the DSD loss (Eq. 5). We use dynamic diffusion timestep that samples larger timestep from range $[0.5,0.75]$ when applying the concept prior and samples smaller timestep from range $[0.1,0.25]$ for the reference prior. We set $\alpha$ as $0.2$ and $\beta$ as $1.0$. The total optimization step is $5000$. We also adopt the total variation loss~\cite{rudin1992nonlinear} and contextual loss~\cite{mechrez2018contextual} to enhance the texture quality. Specially, the contextual loss is applied between the rendered color image and the 20 augmented views of the initial model. The whole 3D-to-3D generation process takes around 2 hours using a single NVIDIA A100 GPU.

\section{User study settings}
We randomly select 20 models from our dataset and generate 3 variations for each model. We invite a total of 30 users, recruited publicly, to complete a questionnaire consisting of 30 pairwise comparisons (15 for image-to-3D and 15 for 3D-to-3D) in person, totaling 900 answers. For image-to-3D, we show two generated 3D models (one by our method and one by the baseline method) beside a concept image and ask the users to answer the question: ``Which of the two models do you prefer (\eg~higher quality and more details) on the premise of aligning with the input view?" For 3D-to-3D, we show two sets of generated 3D variations beside a reference model and ask the question: ``Which of the two sets do you prefer (\eg~higher quality and more diversity) on the premise of sharing consistent themes with the reference?"

\begin{table}
\setlength\tabcolsep{3pt}
  \caption{\revision{Quantitative evaluation of theme-driven diffusion model.}}
  \label{tab:finetune_iteration}
  \footnotesize
  \resizebox{\linewidth}{!}{
  \begin{tabular}{lcccc}
    \toprule
     & Iteration100 & Iteration200 & Iteration300 & Iteration400
    \\
    \midrule
     LPIPS-diversity $\uparrow$ & 0.627 & 0.617 & 0.403 & 0.347
    \\
    \midrule
    LAION-aesthetic-score $\uparrow$ & 6.262 & 6.355 & 6.367 & 5.941
    \\
  \bottomrule
\end{tabular}}
\end{table}

\revision{
\section{Evaluation of theme-driven diffusion model}
To evaluate the influence of different fine-tuning iterations for the theme-driven diffusion model that generates concept images in the first stage, we conduct ablation studies on four settings,~\ie~fine-tuning the theme-driven diffusion model given one 3D exemplar for 100, 200, 300 and 400 iterations.
We use LPIPS-diversity (LPIPS differences across generated images) and LAION-aesthetic-score to estimate the diversity and quality of generated concept images under different settings. The quantitative results are shown in~\reftab{finetune_iteration}. As can be seen, diversity significantly drops when iteration is 300, and quality drops when iteration is 400, both caused by overfitting. We thus set the fine-tuning iteration to 200 for a single exemplar (\refsec{implementation}).
}

\section{Potential ethics issues}
As a generative model, ~\textit{ThemeStation} may pose ethical issues if used to create baleful and fake content, which requires more vigilance and care. We can adopt the commonly used safety checker in existing text-to-image diffusion models to filter out maliciously generated concept images in our first stage to alleviate the potential ethics issues.

\clearpage

\end{document}